\documentclass{bmvc2k}

\usepackage{graphicx}
\usepackage{tikz}
\usetikzlibrary{spy}
\usetikzlibrary{positioning}
\usepackage{adjustbox}

\usepackage{wrapfig}

\usepackage{amsmath,amssymb}

\newcommand{\bm}[1]{\mathbf{#1}} 

\newcommand\raiseT[2]{%
\setbox0\hbox{$#1{#2}$}\raise\dp0\box0}

\usepackage{amsthm}
\newtheorem{lem}{Lemma}
\newtheorem{prop}{Proposition}
\theoremstyle{definition}

\usepackage{booktabs}
\usepackage{xcolor}
\usepackage{multirow}

\title{Flexible Graph Convolutional Network for 3D Human Pose Estimation}

\addauthor{Abu Taib Mohammed Shahjahan}{a_shahj@live.concordia.ca}{1}
\addauthor{A. Ben Hamza}{hamza@ciise.concordia.ca}{1}

\addinstitution{
 Concordia University \\
 Montreal, QC, Canada
}

\runninghead{Shahjahan, Hamza}{Flexible GCN for 3D Human Pose Estimation}

\def\etal{\emph{et al}\bmvaOneDot}

\begin{document}

\maketitle

\begin{abstract}
Although graph convolutional networks exhibit promising performance in 3D human pose estimation, their reliance on one-hop neighbors limits their ability to capture high-order dependencies among body joints, crucial for mitigating uncertainty arising from occlusion or depth ambiguity. To tackle this limitation, we introduce Flex-GCN, a flexible graph convolutional network designed to learn graph representations that capture broader global information and dependencies. At its core is the flexible graph convolution, which aggregates features from both immediate and second-order neighbors of each node, while maintaining the same time and memory complexity as the standard convolution. Our network architecture comprises residual blocks of flexible graph convolutional layers, as well as a global response normalization layer for global feature aggregation, normalization and calibration. Quantitative and qualitative results demonstrate the effectiveness of our model, achieving competitive performance on benchmark datasets. Code is available at: \textcolor{blue}{https://github.com/shahjahan0275/Flex-GCN}
\end{abstract}


\section{Introduction}
The objective of 3D human pose estimation is to predict the 3D positions of body joints from images or videos. This task is essential for interpreting human movements and actions in various computer vision applications, including sports performance analytics and pedestrian behavior analysis~\cite{zhao2019accurate}. For instance, accurately identifying skeletal joints is crucial for assessing sports activities, as it enables a meaningful evaluation of athletes' performance.

Existing approaches to 3D human pose estimation can generally be categorized into one- and two-stage methods. One-stage approaches, also known as direct regression techniques, aim to predict 3D joint locations directly from input images or video frames without intermediary steps. However, these methods often face depth ambiguity, where multiple plausible 3D poses can explain the same 2D observations. They also struggle with complex poses and occlusions~\cite{park20163d,sun2018integral,pavlakos2017coarse,sun2017compositional}. On the other hand, two-stage approaches, also known as 2D-to-3D lifting methods, typically consist of separate stages for joint detection and pose regression. The first stage detects 2D joint positions in the image, and the second stage uses these 2D detections to estimate the 3D joint positions. By incorporating an intermediate step for 2D joint detection, two-stage methods can mitigate challenges such as occlusions and depth ambiguity, resulting in more robust 3D pose estimations compared to their one-stage counterparts. Moreover, they allow for the use of different 2D pose detectors and lifting networks, providing flexibility in designing and optimizing each component separately, thereby potentially leading to higher accuracy~\cite{martinez2017simple, ci2019optimizing,wu20203d,liu2020learning}.

Graph convolutional network (GCN)-based methods have recently demonstrated considerable promise in 3D human pose estimation~\cite{zhao2019semantic,liu2020comprehensive,zou2021modulated,zou2021compositional,zhang2022group}, leveraging the inherent graph structure of the human body, where joints serve as nodes interconnected by edges representing skeletal connections. By capitalizing on this representation, GCN-based models can effectively capture spatial dependencies crucial for accurate pose estimation. While these methods have demonstrated effectiveness in capturing dependencies between body joints, they are, however, inherently limited in their ability to model interactions beyond immediate neighbors. To overcome this challenge, recent approaches have introduced high-order graph convolutions~\cite{zou2020high,quan2021higher,lee2022multi}, which enable information propagation through multiple hops in the graph, allowing the model to gather insights from not only immediate neighbors but also nodes located farther away. This enhances the model's capacity to capture global context and complex relationships between body joints. Another limitation of GCN-based methods is their inherent reliance on the adjacency matrix, which represents the connectivity between body joints in a graph, with non-zero entries indicating the presence of connections between neighboring joints. By modulating the adjacency matrix~\cite{zou2021modulated}, we can incorporate additional information from nodes that are further apart in the graph, allowing the model to capture more complex dependencies and contextual cues.

To address the aforementioned limitations, we propose a novel graph convolutional network, dubbed Flex-GCN, which employs multi-hop neighbors through a flexible scaling factor that controls the balance between the information from immediate neighbors and the information from nodes that are at most two edges away in the graph. In addition to integrating an initial residual connection into the update rule of Flex-GCN, we also modulate the adjacency matrix to enable our model to consider not only the immediate connections between neighboring joints, but also the spatial relationships between distant joints that may contribute to the overall pose configuration. Our contributions are summarized as follows:

\begin{itemize}
\item We present a flexible graph convolutional network (Flex-GCN), which captures high-order dependencies essential for reducing uncertainty due to occlusion or depth ambiguity in 3D human pose estimation. We also theoretically demonstrate the training stability of Flex-GCN.
\item We design a network architecture that includes flexible graph convolutional layers and a global response normalization layer.
\item Experimental results and ablation studies demonstrate the competitive performance of Flex-GCN against strong baselines on two benchmark datasets.
\end{itemize}

\section{Related Work}
\noindent\textbf{3D human pose estimation} aims to estimate the 3D coordinates of the joints in the human body from images or videos. One-stage and two-stage approaches are two common strategies employed in this task. One-stage methods directly regress the 3D pose from input images~\cite{park20163d,sun2018integral,pavlakos2017coarse,sun2017compositional}, while two-stage methods first predict intermediate representations, such as 2D joint locations, before lifting them to 3D space~\cite{martinez2017simple,ci2019optimizing,wu20203d,liu2020learning}. Two-stage methods, often combined with robust 2D joint detectors, typically exhibit better performance, particularly in addressing depth ambiguity challenges. Our proposed method falls under the category of two-stage approaches, with a network architecture design inspired by the ConvNeXt V2 framework~\cite{woo2023convnext}, which leverages a global response normalization layer.

\medskip\noindent\textbf{GCN-based methods for 3D human pose estimation} offer an intuitive paradigm by representing the human body as a graph structure~\cite{zhao2019semantic,liu2020comprehensive,zou2021modulated,zou2021compositional,zhang2022group}, where the joints of the body serve as nodes and the connections between them represent the bones. This approach leverages the inherent spatial relationships between body parts, allowing for the modeling of complex human poses through graph-based representations. Also, by analyzing the connectivity patterns within the skeletal graph, GCN-based methods can infer the positions of individual joints based on information propagated from neighboring nodes, facilitating robust and contextually informed pose predictions. For instance, SemGCN~\cite{zhao2019semantic} integrates semantic information into the graph convolution, allowing the model to combine structural information from the graph with semantic features derived from the data.  In GroupGCN~\cite{zhang2022group}, convolutional operations are performed within distinct groups, each of  which has its own weight matrix and spatial aggregation kernel. Weight Unsharing~\cite{liu2020comprehensive} analyzes the trade-offs between weight sharing and unsharing in GCNs. Modulated GCN~\cite{zou2021modulated} combines weight modulation to learn unique modulation vectors for individual nodes and adjacency modulation to account for additional edges beyond the human skeleton connections. One major limitation of the standard GCN architecture is that it typically operates with one-hop neighbors, which can restrict the ability of GCN-based methods to capture long-range dependencies and complex interactions within the graph. In other words, these methods provide a relatively local perspective of the graph structure, potentially overlooking long-range interactions and intricate dependencies present in human body movements. To mitigate this issue, High-order GCN~\cite{zou2020high} incorporates high-order dependencies among body joints by considering neighbors located multiple hops away during the update of joint features. Similarly, multi-hop Modulated GCN (MM-GCN)~\cite{lee2022multi} involves modulating and fusing features from multi-hop neighbors. Our proposed model differs from these GCN-based approaches in that we employ a new update rule for graph node feature propagation that seamlessly integrates both first- and second-order neighboring information, combined with an initial residual connection, with the aim of learning graph representations that capture more global information and dependencies, while maintaining the time and memory complexity of the standard GCN. We also leverage adjacency modulation to learn additional connections between body joints.

\section{Method}

\subsection{Preliminaries and Problem Statement}
Let $\mathcal{G}=(\mathcal{V},\mathcal{E},\bm{X})$ be an attributed graph, where $\mathcal{V}=\{1,\ldots,N\}$ is the set of $N$ nodes and $\mathcal{E}\subseteq \mathcal{V}\times\mathcal{V}$ is the set of edges, and $\bm{X}$ an $N\times F$ feature matrix of node attributes. We denote by $\bm{A}$ an $N\times N$ adjacency matrix whose $(i,j)$-th entry is equal to 1 if $i$ and $j$ are neighboring nodes, and 0 otherwise. We also denote by $\hat{\bm{A}}=\bm{D}^{-\frac{1}{2}}\bm{A}\bm{D}^{-\frac{1}{2}}$ the normalized adjacency matrix, where $\bm{D}=\mathsf{diag}(\bm{A}\bm{1})$ is the diagonal degree matrix and $\bm{1}$ is a vector of all ones.

Let $\mathcal{D}=\left\{\left(\mathbf{x}_{i}, \mathbf{y}_{i}\right)\right\}_{i=1}^{N}$ be a training set consisting of 2D joint positions $\bm{x}_{i}\in\mathcal{X}\subset\mathbb{R}^2$ and their associated ground-truth 3D joint positions $\bm{y}_{i}\in\mathcal{Y}\subset\mathbb{R}^3$. The aim of 3D human pose estimation is to learn the parameters $\bm{w}$ of a regression model $f_\bm{w}: \mathcal{X} \rightarrow \mathcal{Y}$ by finding a minimizer of the following objective function
\begin{equation}	
\bm{w}^{*}=\arg\min_{\bm{w}}\frac{1}{N}\sum_{i=1}^{N}l(f_\bm{w}(\bm{x}_{i}),\bm{y}_{i}),
\end{equation}
where $l(f_\bm{w}(\bm{x}_{i}),\bm{y}_{i})$ is an empirical regression loss function.

\subsection{Flexible Graph Convolutional Network}
Central to graph neural networks lies the fundamental concept of the feature propagation rule, which determines how information is transmitted among nodes in a graph. This rule entails updating node features by aggregating information from nearby and/or distant neighbors, followed by non-linear activation, to generate an updated node representation. To this end, we propose a flexible graph convolutional network (Flex-GCN) with the following layer-wise update rule for node feature propagation:
\begin{equation}
\bm{H}^{(\ell+1)}=\sigma\Bigl(((1-s)\bm{I}+s\hat{\bm{A}})\hat{\bm{A}}\bm{H}^{(\ell)}\bm{W}^{(\ell)}
+\bm{X}\widetilde{\bm{W}}^{(\ell)}\Bigr),\qquad \ell=0,\dots,L-1,
\label{Eq:IS}
\end{equation}
where $s\in (0,1)$ is a positive scaling parameter, $\bm{W}^{(\ell)}$ and $\widetilde{\bm{W}}^{(\ell)}$ are learnable weight matrices, $\sigma(\cdot)$ is an element-wise activation function, $\bm{H}^{(\ell)}\in\mathbb{R}^{N\times F_{\ell}}$ is the input feature matrix of the $\ell$-th layer with $F_{\ell}$ feature maps. The input of the first layer is  $\bm{H}^{(0)}=\bm{X}$.

The update rule of Flex-GCN is essentially comprised of three main components: (i) feature propagation that combines the features of the 1- and 2-hop neighbors of nodes (i.e., it aggregates information from immediate and high-order neighboring nodes), (ii) feature transformation that applies learnable weight matrices to the node representations to learn an efficient representation of the graph, and (iii) residual connection for ensuring that information from the initial feature matrix is preserved. The initial residual connection used in the proposed model allows information from the initial feature matrix to bypass the current layer and be directly added to the output of the current layer. This helps preserve important information that may be lost during the aggregation process, thereby improving the flow of information through the network. Note that the propagation operation/matrix $\bm{P}=((1-s)\bm{I}+s\hat{\bm{A}})\hat{\bm{A}}$ of our model is a weighted combination of the normalized adjacency matrix and its square. It allows Flex-GCN to capture information from nodes that are not only directly connected (1-hop), but also incorporates information from the neighbors of the neighbors (2-hop). The parameter $s$ helps control the balance between the information from immediate neighbors and the information from nodes that are at most two edges away in the graph. This is particularly valuable for learning graph representations that capture more global information and dependencies.

\medskip\noindent\textbf{Model Complexity.} \quad For simplicity, we assume the feature dimensions are the same across all layers, i.e., $F_{\ell}=F$ for all $\ell$, with $F \ll N$. Multiplying the propagation matrix $((1-s)\bm{I}+s\hat{\bm{A}})\hat{\bm{A}}$ with an embedding $\bm{H}^{(\ell)}$ costs $\mathcal{O}(\Vert\hat{\bm{A}}\Vert_{0}F)$ in time, where $\Vert\hat{\bm{A}}\Vert_{0}$ denotes the number of non-zero entries of the sparse matrix $\hat{\bm{A}}$ (i.e., number of edges in the graph). Multiplying an embedding with a weight matrix costs $\mathcal{O}(NF^2)$. Also, multiplying the initial feature matrix by the residual connection weight matrix costs $\mathcal{O}(NF^2)$. Hence, the time complexity of an $L$-layer Flex-GCN is $\mathcal{O}(L\Vert\hat{\bm{A}}\Vert_{0}F+LNF^2)$. For memory complexity, an $L$-layer Flex-GCN requires $\mathcal{O}(LNF+LF^2)$ in memory, where $\mathcal{O}(LNF)$ is for storing all embeddings and $\mathcal{O}(LF^2)$ is for storing all layer-wise weight matrices. Therefore, our proposed Flex-GCN model has the same time and memory complexity as that of GCN, albeit Flex-GCN takes into account both immediate and distant graph nodes for improved learned node representations. It is important to note that there is no need to explicitly compute the square of the normalized adjacency matrix in the Flex-GCN model. Instead, we perform right-to-left multiplication of the normalized adjacency matrix with the embedding. This process avoids the computational cost associated with matrix exponentiation and simplifies the computation, making our model more efficient while achieving its objectives.

\medskip\noindent\textbf{Numerical Stability.} \quad To demonstrate the numerical stability of the proposed Flex-GCN model, we start with a useful result in matrix analysis~\cite{Riesz1990FA}, which states that the spectral radius of the sum of two commuting matrices is bounded by the sum of the spectral radii of the individual matrices.

\begin{lem}
If two matrices $\bm{M}_{1}$ and $\bm{M}_{2}$ commute, i.e., $\bm{M}_{1}\bm{M}_{2}=\bm{M}_{2}\bm{M}_{1}$, then
\begin{equation}
\rho(\bm{M}_{1}+\bm{M}_{2})\le \rho(\bm{M}_{1}) + \rho(\bm{M}_{2}),
\end{equation}
where $\rho(\cdot)$ denotes matrix spectral radius (i.e., largest absolute value of all eigenvalues).
\end{lem}

Since the eigenvalues of the normalized Laplacian matrix $\bm{L}=\bm{I}-\hat{\bm{A}}$ lie in the interval $[0,2]$, it follows that $\rho(\hat{\bm{A}})\le 1$. Hence, we have the following result, which demonstrates the training stability of the proposed model, with information smoothly propagating through the graph layers without amplifying or dampening effects that could lead to instability.

\begin{prop}
The update rule of Flex-GCN is numerically stable.
\end{prop}
\noindent\textit{Proof.}\quad Recall that the propagation matrix of Flex-GCN is given by
\begin{equation}
\bm{P}=((1-s)\bm{I}+s\hat{\bm{A}})\hat{\bm{A}}=(1-s)\hat{\bm{A}}+s\hat{\bm{A}}^2.
\end{equation}
Since the matrices $(1-s)\hat{\bm{A}}$ and $s\hat{\bm{A}}^2$ satisfy the assumptions of Lemma 1, we have
\begin{equation}
\rho((1-s)\hat{\bm{A}}+s\hat{\bm{A}}^2)\le \rho((1-s)\hat{\bm{A}}) + \rho(s\hat{\bm{A}}^2)\le 1,
\end{equation}
because both $\rho(\hat{\bm{A}})$ and $\rho(\hat{\bm{A}}^2)$ are bounded by 1. Hence, the spectral radius of the propagation matrix is bounded by 1. Consequently, repeated layer-wise application of this propagation operator is stable.

\medskip\noindent\textbf{Adjacency Modulation.}\quad We modulate the normalized adjacency matrix to capture not just the interactions between adjacent nodes, but also the relationships between distant nodes beyond the natural connections of body joints~\cite{zou2021modulated}, yielding a modulated adjacency matrix $\check{\bm{A}}=\hat{\bm{A}}+\bm{Q}$, where $\bm{Q}\in\mathbb{R}^{N\times N}$ is a learnable adjacency modulation matrix. Since the skeleton graph exhibits symmetry, we ensure symmetry in the learnable adjacency modulation matrix by averaging it with its transpose.

\begin{figure}[!htb]
\centering
\includegraphics[scale=.4]{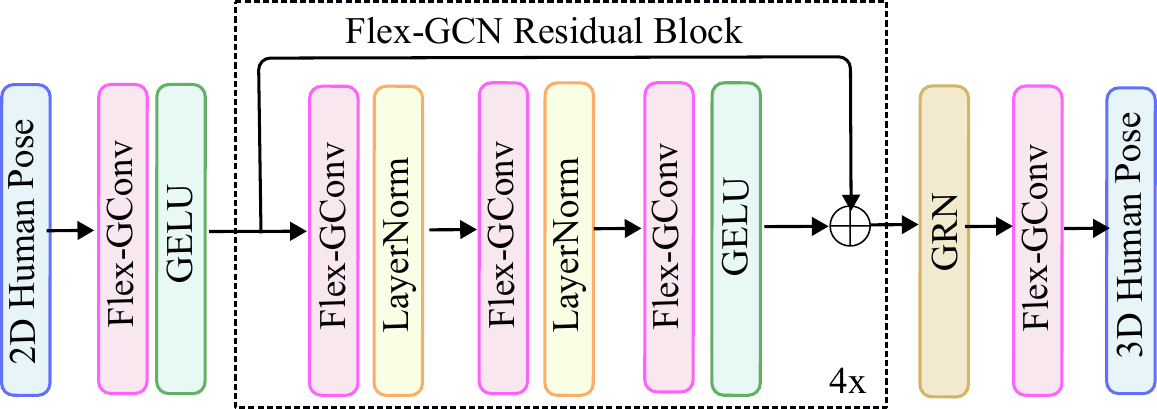}
\caption{Network architecture of Flex-GCN for 3D human pose estimation.}

\label{Fig:network_architecture}
\end{figure}

\medskip\noindent\textbf{Model Architecture.}\quad The overall architecture of our proposed Flex-GCN model is illustrated in Figure~\ref{Fig:network_architecture}. The input to the model is a 2D pose, typically obtained via an off-the-shelf 2D detector ~\cite{chen2018cascaded}, which subsequently undergoes a flexible graph convolutional layer, followed by a GELU activation function. Following the architectural design of the ConvNeXt V2 block ~\cite{Liu2022convnext}, our residual block consists of three flexible graph convolutional (Flex-GConv) layers. In each block, the first two convolutional layers are followed by layer normalization, while the third one is followed by GELU. This residual block is repeated four times. Then, a global response normalization (GRN) layer ~\cite{woo2023convnext} is applied after the residual blocks with the aim of increasing the contrast and selectivity of channels. Finally, the last flexible graph convolutional layer of the network generates the 3D pose.

\medskip\noindent\textbf{Model Prediction.}\quad The output of the last flexible graph convolutional layer contains the final output node embeddings, $\hat{\bm{y}}_{i},\, i=1,\dots,N$, which are the predicted 3D pose coordinates.

\medskip\noindent\textbf{Model Training.}\quad The parameters (i.e., weight matrices for different layers) of Flex-GCN are learned by minimizing the following loss function

\begin{equation}
\mathcal{L} =\frac{1}{N}\left[(1-\alpha)\sum_{i=1}^{N}\Vert\bm{y}_{i}-\hat{\bm{y}}_{i}\Vert_{2}^{2} + \alpha\sum_{i=1}^{N}\Vert\bm{y}_{i}-\hat{\bm{y}}_{i}\Vert_{1}\right],
\end{equation}
which is a weighted sum of the mean squared and mean absolute errors between the 3D ground truth coordinates $\bm{y}_{i}$ and estimated 3D
joint coordinates $\hat{\bm{y}}_{i}$ over $N$ training body joints. The weighting factor $\alpha\in [0,1]$ controls the contribution of each error term.

\section{Experiments}

\subsection{Experimental Setup}
\noindent\textbf{Datasets.}\quad We assess Flex-GCN performance on Human3.6M~\cite{ionescu2013human3} and MPI-INF-3DHP~\cite{Dushyant:2017}.

\medskip\noindent\textbf{Evaluation Protocols and Metrics.}\quad We employ two standard evaluation protocols for training and testing on Human 3.6M designated as Protocol \#1 and Protocol \#2 ~\cite{martinez2017simple}, with associated metrics mean per-joint position error (MPJPE) and Procrustes-aligned mean per-joint position error (PA-MPJPE), respectively. For MPI-INF-3DHP, we use the Area Under Curve (AUC) and Percentage of Correct Keypoint (PCK) as assessment metrics.

\medskip\noindent\textbf{Baselines.}\quad We compare the performance of our model with several state-of-the-art methods for 3D pose estimation, including Weight Unsharing~\cite{liu2020comprehensive}, High-order GCN~\cite{zou2020high}, Pose Grammar and and Data Augmentation~\cite{xu2021monocular}, Compositional GCN (CompGCN)~\cite{zou2021compositional}, Higher-Order Implicit Fairing (HOIF-Net)~\cite{quan2021higher}, Multi-hop Modulated GCN (MM-GCN)~\cite{lee2022multi}, Group GCN~\cite{zhang2022group}, and Modulated GCN~\cite{zou2021modulated}.

\medskip\noindent\textbf{Implementation Details.}\quad All experiments are performed on a Linux machine with a single NVIDIA GeForce RTX A4500 GPU featuring 20GB of memory. We use PyTorch to implement our model and train it for 30 epochs using AMSGrad optimizer on detected 2D poses~\cite{chen2018cascaded}, as well as on ground truth 2D poses. The initial learning rate is set to 0.001, with a decay factor of 0.99 every four epochs, a batch size of 512, and $F=384$. The scaling parameter $s=0.2$ and the weighting factor $\alpha=0.03$ are determined via grid search. To prevent overfitting, we apply dropout with a factor of 0.2 after each graph convolution layer.

\subsection{Results and Analysis}
\noindent\textbf{Quantitative Results on Human3.6M.}\quad In Tables~\ref{Tab:MPJPE_Result} and ~\ref{Tab:PA_MPJPE_Result}, we report the performance comparison of our Flex-GCN model and strong baselines for 3D pose estimation on Human3.6M using the detected 2D pose as input. In both tables, we present the results for all 15 actions, along with their average performance. Table~\ref{Tab:MPJPE_Result} shows that Flex-GCN yields competitive performance, with average MPJPE and PA-MPJPE errors of 46.9mm and 38.6mm, respectively. Under Protocol \#1, Table~\ref{Tab:MPJPE_Result} reveals that Flex-GCN performs better than Modulated GCN~\cite{zou2021modulated} in 14 out of 15 actions, yielding 2.5mm error reduction on average, improving upon this best performing baseline by a relative improvement of 5.08\%. Our model achieves better predictions than the best baseline on challenging actions like hard poses involving activities of self-occlusion such as ``Eating'', ``Sitting'' and ``Smoking'', showing relative error reductions of 1.53\%, 7.47\% and 8.24\%, respectively, in terms of MPJPE. The presence of self-occlusions during activities can pose challenges for human pose estimation, as they restrict the model's access to observable information. For instance, some activities like eating or smoking can lead to occlusions where a person's hands and arms obstruct parts of their face and upper body besides, when sitting, a person's legs and arms may obstruct other body parts such as the torso or feet.

\begin{table*}[!htb]
\caption{Comparison of our model and baseline methods in terms of MPJPE in millimeters, computed between the ground truth and estimated poses on Human3.6M under Protocol \#1. The last column displays the average errors, with boldface numbers denoting the best performance and underlined numbers indicating the second-best performance.}
\footnotesize
\setlength{\tabcolsep}{.8pt}
\smallskip
\centering
\begin{tabular}{l*{17}{c}}
\toprule
& \multicolumn{15}{c}{Action}\\
\cmidrule(lr){2-16}
Method & Dire. & Disc. &  Eat & Greet & Phone & Photo &  Pose & Purch. & Sit & SitD. & Smoke & Wait & WalkD. & Walk & WalkT. & Avg.\\
\midrule
Liu \etal~\cite{liu2020comprehensive} & 46.3 & 52.2 & 47.3 & 50.7 & 55.5 & 67.1 & 49.2 & \underline{46.0} & 60.4 & 71.1 & 51.5 & 50.1 & 54.5 & 40.3 & 43.7 & 52.4\\
Zou \etal~\cite{zou2020high} & 49.0& 54.5& 52.3& 53.6& 59.2 &71.6& 49.6& 49.8 &66.0 &75.5 &55.1 &53.8& 58.5& 40.9 & 45.4 &55.6\\
Xu \etal~\cite{xu2021monocular} & 47.1 & 52.8 & 54.2 & 54.9 & 63.8 & 72.5 & 51.7 & 54.3 & 70.9 & 85.0 & 58.7 & 54.9 & 59.7 & 43.8 & 47.1 & 58.1\\
Zou \etal~\cite{zou2021compositional} & 48.4 & 53.6 & 49.6 & 53.6 & 57.3 & 70.6 & 51.8 & 50.7 & 62.8 & 74.1 & 54.1 & 52.6 & 58.2 & 41.5 & 45.0 & 54.9\\
Quan \etal~\cite{quan2021higher} & 47.0 & 53.7 & 50.9 & 52.4 & 57.8 & 71.3 & 50.2 & 49.1 & 63.5 & 76.3 & 54.1 & 51.6 & 56.5 & 41.7 & 45.3 & 54.8\\
Zou \etal~\cite{zou2021modulated} & 45.4 & \underline{49.2} & \underline{45.7} & \underline{49.4} & \underline{50.4} & \underline{58.2} & \underline{47.9} & \underline{46.0} & \underline{57.5} & \textbf{63.0} & \underline{49.7} & \underline{46.6} & \underline{52.2} & \underline{38.9} & \underline{40.8} & \underline{49.4}\\
Lee \etal~\cite{lee2022multi} & 46.8 & 51.4 & 46.7 & 51.4 & 52.5 & 59.7 & 50.4 & 48.1 & 58.0 & 67.7 & 51.5 & 48.6 & 54.9 & 40.5 & 42.2 & 51.7\\
Zhang \etal~\cite{zhang2022group} & \underline{45.0} & 50.9 & 49.0 & 49.8 & 52.2 & 60.9 & 49.1 & 46.8 & 61.2 & 70.2 & 51.8 & 48.6 & 54.6 & 39.6 & 41.2 & 51.6\\
\midrule
Ours & \textbf{40.2} & \textbf{45.8} & \textbf{45.0} & \textbf{46.8} &\textbf{48.6} & \textbf{54.0} & \textbf{42.4} & \textbf{42.1} & \textbf{53.2} & \underline{66.7}  &\textbf{45.6} & \textbf{45.4} & \textbf{48.8} & \textbf{38.4} & \textbf{40.1} & \textbf{46.9} \\

\bottomrule
\end{tabular}
\label{Tab:MPJPE_Result}
\end{table*}

Under Protocol \#2, Table~\ref{Tab:PA_MPJPE_Result} shows that our model on average reduces the error by 1.28\% compared to Modulated GCN~\cite{zou2021modulated}, and achieves better results in 12 out of 15 actions, with same performance in the action 'phone'. Also, our method outperforms Modulated GCN on the challenging actions of ``Greeting'', ``Sitting'' and ``Smoking'', yielding relative error reductions of 2\%, 4.74\% and 5.67\%, respectively, in terms of PA-MPJPE. Moreover, our model performs better than Modulated GCN on the challenging ``Photo'' action, yielding a relative error reduction of 2\%. In addition, Flex-GCN outperforms High-order GCN~\cite{zou2020high} by a relative improvement of 11.67\% on average, as well as on all actions.

\begin{table*}[!htb]
\caption{Comparison of our model and baseline methods in terms of PA-MPJPE, computed between the ground truth and estimated poses on Human3.6M under Protocol \#2.}
\footnotesize
\setlength{\tabcolsep}{.8pt}
\smallskip
\centering
\begin{tabular}{l*{17}{c}}
\toprule
& \multicolumn{15}{c}{Action}\\
\cmidrule(lr){2-16}
Method & Dire. & Disc. &  Eat & Greet & Phone & Photo &  Pose & Purch. & Sit & SitD. & Smoke & Wait & WalkD. & Walk & WalkT. & Avg.\\
\midrule
Liu \etal~\cite{liu2020comprehensive} & 35.9 & 40.0 & 38.0 & 41.5 & 42.5 & 51.4 &  37.8 & 36.0 & 48.6 & 56.6 & 41.8 & 38.3 & 42.7 & 31.7 & 36.2 & 41.2\\
Zou \etal~\cite{zou2020high} &38.6 &42.8& 41.8 &43.4 &44.6& 52.9& 37.5& 38.6 &53.3 &60.0& 44.4& 40.9& 46.9 &32.2 &37.9 &43.7\\
Xu \etal~\cite{xu2021monocular} & 36.7 & 39.5 & 41.5 & 42.6 & 46.9 & 53.5 & 38.2 & 36.5 & 52.1 & 61.5 & 45.0 & 42.7 & 45.2 & 35.3 & 40.2 & 43.8\\
Zou \etal~\cite{zou2021compositional} & 38.4 & 41.1 & 40.6 & 42.8 & 43.5 & 51.6 & 39.5 & 37.6 & 49.7 & 58.1 & 43.2 & 39.2 & 45.2 & 32.8 & 38.1 & 42.8\\
Quan \etal~\cite{quan2021higher} & 36.9 & 42.1 & 40.3 & 42.1 & 43.7 & 52.7 & 37.9 & 37.7 & 51.5 & 60.3 & 43.9 & 39.4 & 45.4 & 31.9 & 37.8 & 42.9\\
Zou \etal~\cite{zou2021modulated} & 35.7 & \underline{38.6} & \textbf{36.3} & \underline{40.5} & \textbf{39.2} & \underline{44.5} & 37.0 & 35.4 & \underline{46.4} & \textbf{51.2} & \underline{40.5} & \textbf{35.6} & \underline{41.7} & \underline{30.7} & 33.9 & \underline{39.1}\\
Lee \etal~\cite{lee2022multi} & 35.7 & 39.6 & 37.3 & 41.4 & 40.0 & 44.9 & 37.6 & 36.1 & 46.5 & \underline{54.1} & 40.9 & 36.4 & 42.8 & 31.7 & 34.7 & 40.3\\
Zhang \etal~\cite{zhang2022group} & \underline{35.3} & 39.3 & 38.4 & 40.8 & 41.4 & 45.7 & \underline{36.9} & \underline{35.1} & 48.9 & 55.2 & 41.2 & 36.3 & 42.6 & 30.9 & \underline{33.7} & 40.1\\
\midrule
Ours & \textbf{34.1} & \textbf{38.0} & \underline{36.8} & \textbf{39.7} &\textbf{39.2} & \textbf{43.6} & \textbf{33.4} & \textbf{34.5} & \textbf{44.2} & 57.1 & \textbf{38.3} & \underline{36.0} & \textbf{41.0} & \textbf{29.9} & \textbf{33.1} &\textbf{38.6}\\
\bottomrule
\end{tabular}
\label{Tab:PA_MPJPE_Result}
\end{table*}

\begin{figure}[!htb]
\centering
\includegraphics[scale=.7]{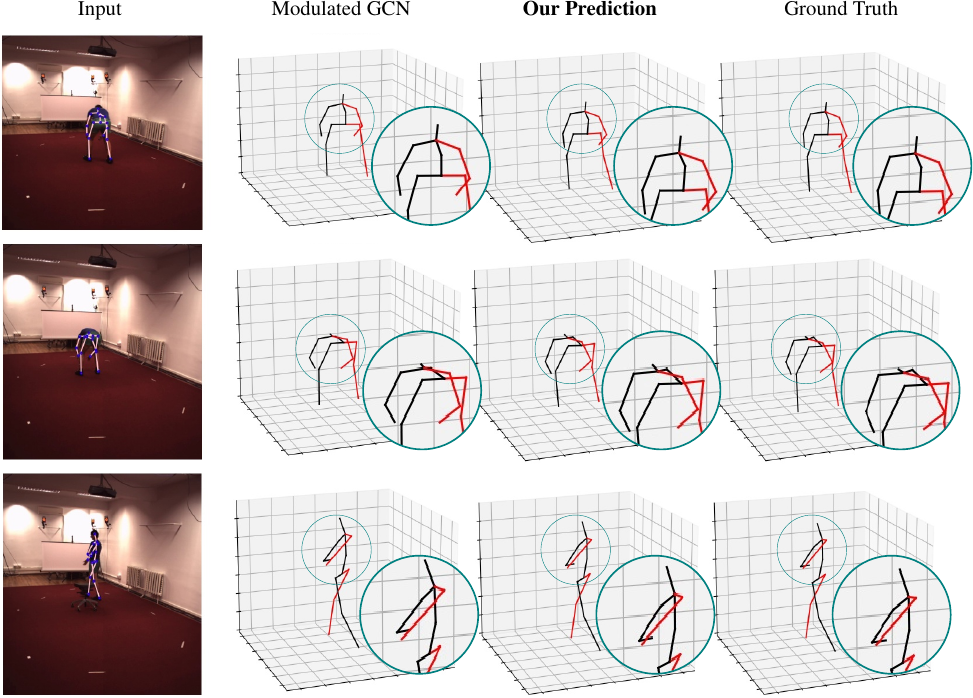}
\caption{Visual comparison between Flex-GCN and Modulated GCN on sample actions from the Human3.6M dataset.}
\label{Fig:Qualitative}
\end{figure}

\medskip\noindent\textbf{Qualitative Results.}\quad Figure~\ref{Fig:Qualitative} displays the visual results obtained by Flex-GCN on sample actions from the Human3.6M dataset. The effectiveness of our model is demonstrated by the close alignment between the predicted 3D poses and the ground truth, as shown in Figure~\ref{Fig:Qualitative}. Compared to Modulated GCN, our model generates poses that closely resemble the ground truth, even in hard poses with self-occlusions.

\medskip

\newlength{\oldintextsep}
\setlength{\oldintextsep}{\intextsep}
\setlength\intextsep{0pt}
\begin{wraptable}{r}{0.4\textwidth}
\small
\setlength{\tabcolsep}{1pt}
\centering
\caption{Results on MPI-INF-3DHP.}
\smallskip
\begin{tabular}{lcc}
\toprule
Method & PCK $(\uparrow)$ & AUC $(\uparrow)$\\
\midrule
Xu \etal~\cite{xu2021graph} & 80.1 & 45.8 \\
Zeng \etal~\cite{zeng2021learning} & \underline{82.1} & 46.2 \\
Lee \etal~\cite{lee2022multi} & 81.6 & \underline{50.3}\\
Zhang \etal~\cite{zhang2022group} & 81.1 & 49.9\\
\midrule
Ours & \textbf{85.2} & \textbf{51.8} \\
\bottomrule
\end{tabular}
\label{Tab:mpi_3dhp_inf}
\end{wraptable}

\medskip\noindent\textbf{Cross-Dataset Results on MPI-INF-3DHP.}\quad In Table~\ref{Tab:mpi_3dhp_inf}, We evaluate the generalization ability of our method by comparing it against strong baselines using a different dataset. Our model is trained on Human3.6M and evaluated on MPI-INF-3DHP. Results demonstrate that our model consistently outperforms the baselines, achieving relative improvements of 1.05\% and 2.9\% in terms of PCK and AUC metrics, respectively, in comparison with the best performing baselines. This highlights the strong generalization capability of our model to unseen scenarios and datasets.

\medskip\noindent\textbf{Ground Truth Results and Runtime Analysis.}\quad Table~\ref{Tab:baselineComparison} (left) shows that our model consistently yields better performance than the baselines under both Protocols \#1 and \#2 on the Human3.6M dataset using the ground truth 2D pose as input. This comparison highlights the efficacy of our model in leveraging ground truth 2D poses to generate more accurate 3D pose estimates. In addition to evaluating the accuracy and effectiveness of Flex-GCN, we conduct an analysis of its inference time to assess the efficiency of our approach in processing and generating outputs. The results of the inference time analysis are presented in Table~\ref{Tab:baselineComparison} (right), where the computational performance of our Flex-GCN model is compared against that of strong baseline methods. Notably, our model demonstrates significantly improved inference time compared to the baseline methods, highlighting its computational efficiency.


\medskip
\begin{table}[!htb]
\caption{\textbf{Left:} Performance comparison of our model and baselines on Human3.6M using the ground truth 2D pose as input. \textbf{Right:} Runtime analysis.}
\small
\setlength{\tabcolsep}{4.5pt}
\smallskip
\centering
\begin{tabular}{cc}
\begin{tabular}{lcc}
\toprule
Method & MPJPE $(\downarrow)$ & PA-MPJPE $(\downarrow)$ \\
\midrule
SemGCN~\cite{zhao2019semantic} & 42.14 & 33.53 \\
High-order GCN~\cite{zou2020high} & 39.52 & 31.07\\
Modulated GCN~\cite{zou2021modulated} & 38.25 & 30.06\\
Weight Unsharing~\cite{liu2020comprehensive} & 37.83 & 30.09\\
\midrule
Ours  & \textbf{37.41} & \textbf{29.87} \\
\bottomrule
\end{tabular} &
\begin{tabular}{lc}
\toprule
Method &  Inference Time \\
\midrule
High-Order GCN~\cite{zou2020high}  & .013s \\
Weight Unsharing~\cite{liu2020comprehensive} & .032s \\
MM-GCN~\cite{lee2022multi} & .009s\\
Modulated GCN~\cite{zou2021modulated} & .010s \\
\midrule
Ours & 0.06s \\
\bottomrule
\end{tabular}
\end{tabular}
\label{Tab:baselineComparison}
\end{table}

\subsection{Ablation Study}
\begin{figure}[!htb]
\centering
\begin{tabular}{cc}
\includegraphics[width=2.4in]{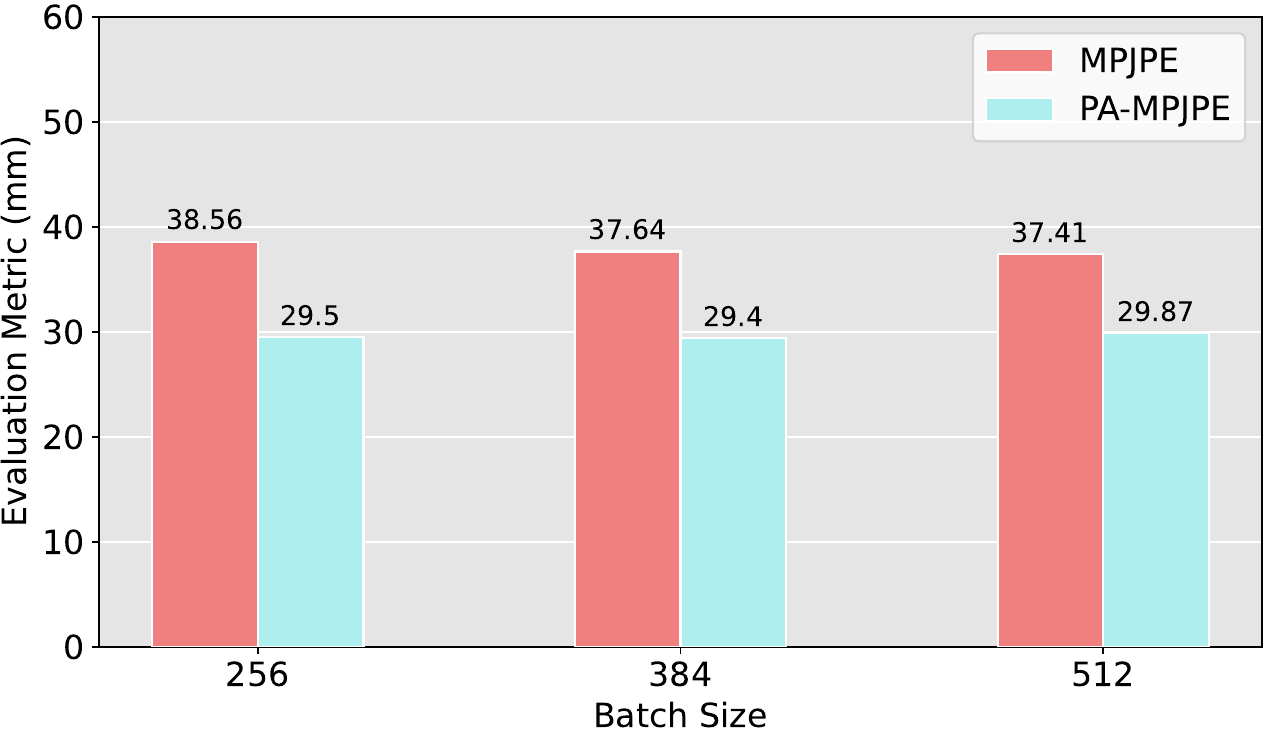} & \includegraphics[width=2.4in]{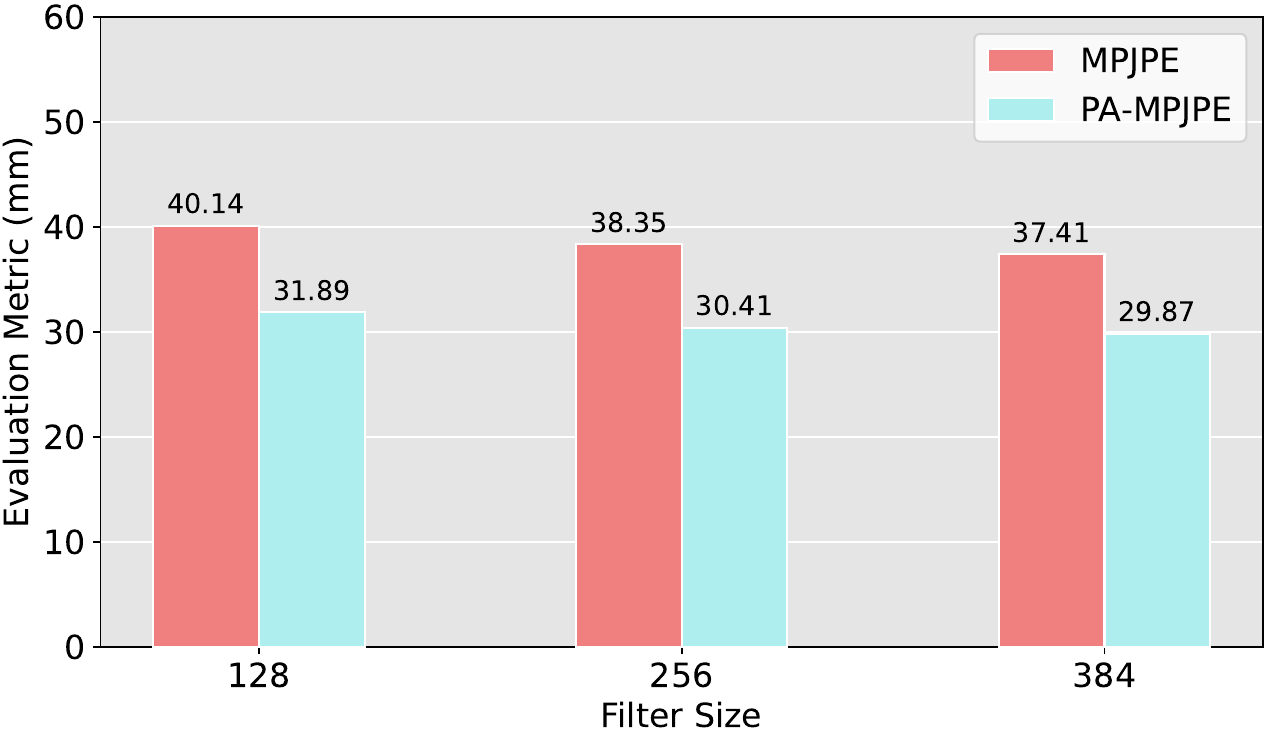}
\end{tabular}
\caption{Performance of our proposed Flex-GCN model on the Human3.6M dataset using varying batch and filter sizes.}
\label{Fig:FigureBatchFilter}
\end{figure}

\noindent\textbf{Effect of Batch and Filter Size.}\quad In Figure~\ref{Fig:FigureBatchFilter}, we analyze the impact of varying batch and filter sizes on our model's performance. This analysis is crucial as both parameters play a significant role in the training efficiency and the overall accuracy of the model. Batch size directly influences the stability and speed of the training process. Smaller batch sizes can lead to noisier gradient estimates but allow for more frequent updates, potentially improving generalization. Larger batch sizes, on the other hand, provide more stable gradient estimates but require more memory and can lead to slower convergence. Filter size, which determines the number of learnable parameters in each layer of the network, affects the model's capacity to capture complex patterns. A larger filter size increases the model's ability to learn intricate features but also raises the risk of overfitting. Conversely, a smaller filter size may lead to underfitting. Our analysis indicates that a batch size of 512 and a filter size of 384 result in the best performance. This combination yields the lowest MPJPE and PA-MPJPE values, respectively, signifying that the model is accurately estimating 3D human poses. The batch size of 512 strikes a balance between gradient stability and update frequency, while the filter size of 384 provides a sufficient number of parameters to learn detailed features without overfitting.

\medskip

\medskip\noindent\textbf{Effect of Residual Connection and Modulation Symmetry.}\quad We analyze the impact of the initial residual connection (IRC) in the layer-wise propagation rule on our model performance, and the results are reported in Table~\ref{Tab:skipConnection} (left). The inclusion of IRC shows significant improvements, with relative error reductions of 6.58\% and 4.41\% in terms of MPJPE and PA-MPJPE, respectively. This demonstrates the effectiveness of IRC in enhancing our model's accuracy. By preserving and reinforcing the initial node features throughout the layers, IRC facilitates more stable and effective learning, ensuring that essential positional information is maintained and utilized in subsequent layers.

Furthermore, we assess the effect of symmetrizing the learnable adjacency modulation matrix on model performance. The results, presented in Table~\ref{Tab:skipConnection} (right), indicate that introducing symmetry into the adjacency modulation process yields tangible reductions in both MPJPE and PA-MPJPE. Specifically, the MPJPE error decreased by 0.58mm and the PA-MPJPE error decreased by 0.24mm, compared to the configuration without symmetry. This improvement highlights the advantage of leveraging skeleton graph symmetry to enhance the precision of our model's estimations. By ensuring that the relational information between joints is consistently balanced, the symmetric adjacency modulation matrix enables more accurate and reliable pose estimations. This approach not only refines the positional accuracy of individual joints but also improves the overall coherence of the estimated poses, demonstrating the critical role of structured graph modulation in 3D human pose estimation.


\medskip
\begin{table}[!htb]
\caption{Effect of initial residual connection (IRC) and symmetry of modulation adjacency.}
\small
\setlength{\tabcolsep}{4pt}
\smallskip
\centering
\begin{tabular}{cc}
\begin{tabular}{lcc}
\toprule
Method & MPJPE $(\downarrow)$ & PA-MPJPE $(\downarrow)$ \\
\midrule
Without IRC & 39.76 & 31.25 \\
With IRC & \textbf{37.41} & \textbf{29.87}\\
\bottomrule
\end{tabular} &
\begin{tabular}{l*{7}{c}}
\toprule
Method & MPJPE $(\downarrow)$  & PA-MPJPE $(\downarrow)$ \\
\midrule
Without Symmetry & 37.99 & 30.11 \\
With Symmetry & \textbf{37.41} & \textbf{29.87}\\
\bottomrule
\end{tabular}
\end{tabular}
\label{Tab:skipConnection}
\end{table}

\section{Conclusion}
We introduced a simple yet efficient Flex-GCN model, which captures high-order dependencies essential for reducing uncertainty due to occlusion or depth ambiguity in 3D human pose estimation. We also theoretically demonstrated the training stability of Flex-GCN. Experimental results demonstrate that our model outperforms competitive baselines on standard datasets for 3D human pose estimation. Furthermore, our exploration of adjacency modulation enables Flex-GCN to incorporate richer contextual information beyond the natural connections of body joints, leading to enhanced performance in challenging scenarios. Through ablation studies, we have elucidated the contributions of various design choices, such as the initial residual connection and symmetry of modulation adjacency, highlighting their positive impact on model performance. For future work, we intend to apply our model to a broader range of computer vision and graph representation learning tasks.

\bigskip\noindent\textbf{Acknowledgments.}\quad This work was supported in part by the Discovery Grants Program of the Natural Sciences and Engineering Research Council of Canada under grant RGPIN-2024-04291.

\bibliography{references}

\begin{thebibliography}{26}
\providecommand{\natexlab}[1]{#1}
\providecommand{\url}[1]{\texttt{#1}}
\expandafter\ifx\csname urlstyle\endcsname\relax
  \providecommand{\doi}[1]{doi: #1}\else
  \providecommand{\doi}{doi: \begingroup \urlstyle{rm}\Url}\fi

\bibitem[Chen et~al.(2018)Chen, Wang, Peng, Zhang, Yu, and
  Sun]{chen2018cascaded}
Yilun Chen, Zhicheng Wang, Yuxiang Peng, Zhiqiang Zhang, Gang Yu, and Jian Sun.
\newblock Cascaded pyramid network for multi-person pose estimation.
\newblock In \emph{Proc. IEEE Conference on Computer Vision and Pattern
  Recognition}, pages 7103--7112, 2018.

\bibitem[Ci et~al.(2019)Ci, Wang, Ma, and Wang]{ci2019optimizing}
Hai Ci, Chunyu Wang, Xiaoxuan Ma, and Yizhou Wang.
\newblock Optimizing network structure for {3D} human pose estimation.
\newblock In \emph{Proc. IEEE International Conference on Computer Vision},
  pages 2262--2271, 2019.

\bibitem[Ionescu et~al.(2013)Ionescu, Papava, Olaru, and
  Sminchisescu]{ionescu2013human3}
Catalin Ionescu, Dragos Papava, Vlad Olaru, and Cristian Sminchisescu.
\newblock {Human3.6M}: Large scale datasets and predictive methods for {3D}
  human sensing in natural environments.
\newblock \emph{IEEE Transactions on Pattern Analysis and Machine
  Intelligence}, 36\penalty0 (7):\penalty0 1325--1339, 2013.

\bibitem[Lee and Kim(2022)]{lee2022multi}
Jae~Yung Lee and I~Gil Kim.
\newblock Multi-hop modulated graph convolutional networks for {3D} human pose
  estimation.
\newblock In \emph{Proc. British Machine Vision Conference}, 2022.

\bibitem[Liu et~al.(2020{\natexlab{a}})Liu, Ding, Zou, Wang, and
  Tang]{liu2020comprehensive}
Kenkun Liu, Rongqi Ding, Zhiming Zou, Le~Wang, and Wei Tang.
\newblock A comprehensive study of weight sharing in graph networks for {3D}
  human pose estimation.
\newblock In \emph{Proc. European Conference on Computer Vision}, pages
  318--334, 2020{\natexlab{a}}.

\bibitem[Liu et~al.(2020{\natexlab{b}})Liu, Zou, and Tang]{liu2020learning}
Kenkun Liu, Zhiming Zou, and Wei Tang.
\newblock Learning global pose features in graph convolutional networks for
  {3D} human pose estimation.
\newblock In \emph{Proc. Asian Conference on Computer Vision},
  2020{\natexlab{b}}.

\bibitem[Liu et~al.(2022)Liu, Mao, Wu, Feichtenhofer, Darrell, and
  Xie]{Liu2022convnext}
Zhuang Liu, Hanzi Mao, Chao-Yuan Wu, Christoph Feichtenhofer, Trevor Darrell,
  and Saining Xie.
\newblock A {ConvNet} for the 2020s.
\newblock In \emph{Proc. IEEE Conference on Computer Vision and Pattern
  Recognition}, pages 11976--11986, 2022.

\bibitem[Martinez et~al.(2017)Martinez, Hossain, Romero, and
  Little]{martinez2017simple}
Julieta Martinez, Rayat Hossain, Javier Romero, and James~J Little.
\newblock A simple yet effective baseline for {3D} human pose estimation.
\newblock In \emph{Proc. IEEE International Conference on Computer Vision},
  pages 2640--2649, 2017.

\bibitem[Mehta et~al.(2017)Mehta, Rhodin, Casas, Fua, Sotnychenko, Xu, and
  Theobalt]{Dushyant:2017}
Dushyant Mehta, Helge Rhodin, Dan Casas, Pascal Fua, Oleksandr Sotnychenko,
  Weipeng Xu, and Christian Theobalt.
\newblock Monocular {3D} human pose estimation in the wild using improved {CNN}
  supervision.
\newblock In \emph{Proc. International Conference on 3D Vision}, 2017.

\bibitem[Park et~al.(2016)Park, Hwang, and Kwak]{park20163d}
Sungheon Park, Jihye Hwang, and Nojun Kwak.
\newblock {3D} human pose estimation using convolutional neural networks with
  {2D} pose information.
\newblock In \emph{Proc. European Conference on Computer Vision}, pages
  156--169. Springer, 2016.

\bibitem[Pavlakos et~al.(2017)Pavlakos, Zhou, Derpanis, and
  Daniilidis]{pavlakos2017coarse}
Georgios Pavlakos, Xiaowei Zhou, Konstantinos~G Derpanis, and Kostas
  Daniilidis.
\newblock Coarse-to-fine volumetric prediction for single-image {3D} human
  pose.
\newblock In \emph{Proc. IEEE Conference on Computer Vision and Pattern
  Recognition}, pages 7025--7034, 2017.

\bibitem[Quan and Hamza(2021)]{quan2021higher}
Jianning Quan and A~Ben Hamza.
\newblock Higher-order implicit fairing networks for {3D} human pose
  estimation.
\newblock In \emph{Proc. British Machine Vision Conference}, 2021.

\bibitem[Riesz and Sz.-Nagy(1990)]{Riesz1990FA}
Frigyes Riesz and Bela Sz.-Nagy.
\newblock \emph{Functional Analysis}.
\newblock Dover Publications, 1990.

\bibitem[Sun et~al.(2017)Sun, Shang, Liang, and Wei]{sun2017compositional}
Xiao Sun, Jiaxiang Shang, Shuang Liang, and Yichen Wei.
\newblock Compositional human pose regression.
\newblock In \emph{Proc. IEEE International Conference on Computer Vision},
  pages 2602--2611, 2017.

\bibitem[Sun et~al.(2018)Sun, Xiao, Wei, Liang, and Wei]{sun2018integral}
Xiao Sun, Bin Xiao, Fangyin Wei, Shuang Liang, and Yichen Wei.
\newblock Integral human pose regression.
\newblock In \emph{Proc. European Conference on Computer Vision}, pages
  529--545, 2018.

\bibitem[Woo et~al.(2023)Woo, Debnath, Hu, Chen, Liu, Kweon, and
  Xie]{woo2023convnext}
Sanghyun Woo, Shoubhik Debnath, Ronghang Hu, Xinlei Chen, Zhuang Liu, In~So
  Kweon, and Saining Xie.
\newblock {ConvNeXt V2}: Co-designing and scaling {ConvNets} with masked
  autoencoders.
\newblock In \emph{Proc. IEEE Conference on Computer Vision and Pattern
  Recognition}, pages 16133--16142, 2023.

\bibitem[Wu and Xiao(2020)]{wu20203d}
Haiping Wu and Bin Xiao.
\newblock {3D} human pose estimation via explicit compositional depth maps.
\newblock In \emph{Proc. AAAI Conference on Artificial Intelligence},
  volume~34, pages 12378--12385, 2020.

\bibitem[Xu and Takano(2021)]{xu2021graph}
Tianhan Xu and Wataru Takano.
\newblock Graph stacked hourglass networks for {3D} human pose estimation.
\newblock In \emph{Proc. IEEE Conference on Computer Vision and Pattern
  Recognition}, pages 16105--16114, 2021.

\bibitem[Xu et~al.(2021)Xu, Wang, Liu, Liu, Xie, and Zhu]{xu2021monocular}
Yuanlu Xu, Wenguan Wang, Tengyu Liu, Xiaobai Liu, Jianwen Xie, and Song-Chun
  Zhu.
\newblock Monocular {3D} pose estimation via pose grammar and data
  augmentation.
\newblock \emph{IEEE Transactions on Pattern Analysis and Machine
  Intelligence}, 2021.

\bibitem[Zeng et~al.(2021)Zeng, Sun, Yang, Zhao, Liu, and Xu]{zeng2021learning}
Ailing Zeng, Xiao Sun, Lei Yang, Nanxuan Zhao, Minhao Liu, and Qiang Xu.
\newblock Learning skeletal graph neural networks for hard {3D} pose
  estimation.
\newblock In \emph{Proc. IEEE International Conference on Computer Vision},
  pages 11436--11445, 2021.

\bibitem[Zhang(2022)]{zhang2022group}
Zijian Zhang.
\newblock Group graph convolutional networks for {3D} human pose estimation.
\newblock In \emph{Proc. British Machine Vision Conference}, 2022.

\bibitem[Zhao et~al.(2019{\natexlab{a}})Zhao, Peng, Tian, Kapadia, and
  Metaxas]{zhao2019semantic}
Long Zhao, Xi~Peng, Yu~Tian, Mubbasir Kapadia, and Dimitris~N Metaxas.
\newblock Semantic graph convolutional networks for {3D} human pose regression.
\newblock In \emph{Proc. IEEE Conference on Computer Vision and Pattern
  Recognition}, pages 3425--3435, 2019{\natexlab{a}}.

\bibitem[Zhao et~al.(2019{\natexlab{b}})Zhao, Yuan, and Chen]{zhao2019accurate}
Yun Zhao, Zejian Yuan, and Badong Chen.
\newblock Accurate pedestrian detection by human pose regression.
\newblock \emph{IEEE Transactions on Image Processing}, 29:\penalty0
  1591--1605, 2019{\natexlab{b}}.

\bibitem[Zou and Tang(2021)]{zou2021modulated}
Zhiming Zou and Wei Tang.
\newblock Modulated graph convolutional network for {3D} human pose estimation.
\newblock In \emph{Proc. IEEE International Conference on Computer Vision},
  pages 11477--11487, 2021.

\bibitem[Zou et~al.(2020)Zou, Liu, Wang, and Tang]{zou2020high}
Zhiming Zou, Kenkun Liu, Le~Wang, and Wei Tang.
\newblock High-order graph convolutional networks for {3D} human pose
  estimation.
\newblock In \emph{Proc. British Machine Vision Conference}, 2020.

\bibitem[Zou et~al.(2021)Zou, Liu, Wu, and Tang]{zou2021compositional}
Zhiming Zou, Tianqi Liu, Dapeng Wu, and Wei Tang.
\newblock Compositional graph convolutional networks for {3D} human pose
  estimation.
\newblock In \emph{Proc. IEEE International Conference on Automatic Face and
  Gesture Recognition}, 2021.

\end{thebibliography}
\end{document}